\documentclass[11pt]{article}
\usepackage{booktabs}
\usepackage{longtable}
\usepackage{amsmath}
\usepackage{booktabs}
\usepackage[preprint]{acl}

\usepackage{times}
\usepackage{latexsym}
\usepackage[utf8]{inputenc}
\usepackage[T2A]{fontenc}
\usepackage[russian,english]{babel}
\usepackage{multirow}
\usepackage[most]{tcolorbox}
\usepackage{xcolor} 
\usepackage{textcomp}
\usepackage[table]{xcolor}
\usepackage[most]{tcolorbox}
\usepackage{multirow}

\usepackage{array}
\usepackage[T1]{fontenc}

\usepackage[utf8]{inputenc}

\usepackage{microtype}
\usepackage{caption}
\usepackage{inconsolata}

\usepackage{graphicx}
\usepackage{url}

\usepackage{fontspec}
\newfontfamily\kalpurush{kalpurush.ttf}

\usepackage{inconsolata}
\usepackage{import}

\usepackage{booktabs}
\usepackage{siunitx}
\usepackage{placeins}

\title{Polite on the Surface, Broken in Practice: A Curated Dataset for Fixing Generation and Register Failures in Low-Resource Bangla Text Generation}

\author{
\textbf{Md. Asaduzzaman Shuvo},
\textbf{Mahedi Hasan},\\
\textbf{Md. Tashin Parvez},
\textbf{Azizul Haque Noman},
\textbf{Md. Shafayet Hossain Ovi},
\\
\\
United International University, Bangladesh 
\\
\small{
 \textbf{Emails:} \texttt{\{ashuvo221104,mhasan221119\}@bscse.uiu.ac.bd};} \\ \small{ \texttt{\{tashinparvez2002,azizulhaquenoman,shafayethossain5463\}@gmail.com};}
}

\begin{document}

\maketitle
\begin{abstract}
Recent advances in Multilingual Large Language Models (MLLMs) have improved cross-lingual conversational capabilities, yet generating reliable, institutionally usable text in low-resource languages like Bangla remains a critical bottleneck, particularly for compact, low-resource LLMs. Beyond surface fluency, state-of-the-art and small open-weight models frequently exhibit deeper generation failures in Bangla: incoherent or off-topic output, structural misformatting, and inconsistent honorific registers, that render text unusable in formal contexts. To address this, we introduce \textbf{BLADE (BangLa Application and DialoguE generation)}, a culturally aligned instruction-tuning dataset and benchmarking framework comprising $4,196$ curated interaction pairs. We use this resource to fine-tune and evaluate compact open-weight architectures, including DeepSeek-8B, LLaMA-3.2-3B, Qwen2-1.5B, and TigerLLM-1B, via LoRA adapters under 4-bit NormalFloat (NF4) quantization on limited hardware. Our evaluations show that targeted fine-tuning on BLADE yields substantial improvements in generation reliability, structural fidelity, and register consistency even for the smallest models, establishing a rigorous benchmark for closing generation and pragmatic gaps in low-resource multilingual text generation. Code and dataset: \url{https://github.com/ashuvo25/Bangla_Application_LLM/tree/main}
\end{abstract}

\section{Introduction}
Large Language Models (LLMs) have fundamentally transformed multilingual text generation \citep{achiam2023gpt,touvron2023llama2openfoundation}, yet for low-resource languages this progress remains superficial, and the gap widens further for \textit{low-resource LLMs}, compact, open-weight models adapted under limited compute and data. The \textit{multilingual curse} manifests not merely as vocabulary gaps, but as a deeper \textit{generation reliability gap}: the divide between surface fluency and functional usability. This gap surfaces as outright generation failures, incoherent, off-topic, or structurally broken output, compounded by subtler pragmatic failures such as inconsistent honorific registers, both of which render text institutionally unusable.

Bangla, the fifth most spoken language worldwide with over 230 million native speakers, makes this failure concrete. Current multilingual models, and especially compact low-resource LLMs commonly deployed for low-resource-language tasks, produce Bangla text ranging from incoherent or off-topic to merely superficially fluent: misplaced document structures, inconsistent honorific registers, degenerate content, and culturally incongruent discourse markers that make outputs institutionally unacceptable \citep{mukherjee-etal-2025-women, snigdha2022representation}.

\begin{table}[t]
\centering
\small
\begin{tabular}{p{0.46\linewidth} p{0.46\linewidth}}
\toprule
\textbf{\color{red!70!black}Zero-Shot (Flawed)} & \textbf{\color{green!60!black}BLADE-SFT (Correct)} \\
\midrule
\begin{tcolorbox}[colback=red!5, colframe=red!20, arc=2pt, boxrule=0.5pt, left=2pt, right=2pt, top=2pt, bottom=2pt]
{\fontspec{kalpurush.ttf}\footnotesize
...প্রাইভা, সাধারণত ১৫-১৭.০২.২০২৩ | শুখ: উচ্চকেন্দ্র বিবর্তন: ঢঙের ঘনমঞ্জর... ভারসাম্য: −২০°সে/−৪°ফা...
}
\end{tcolorbox}
\textit{\small (English: fragmented, off-topic output unrelated to the "leave application" prompt.)}
&
\begin{tcolorbox}[colback=green!5, colframe=green!20, arc=2pt, boxrule=0.5pt, left=2pt, right=2pt, top=2pt, bottom=2pt]
{\fontspec{kalpurush.ttf}\footnotesize
তারিখঃ ১৮/১০/২০২৪ খ্রিঃ ... বিষয়ঃ অগ্রিম ছুটির জন্য প্রধান শিক্ষকের নিকট আবেদন। মহোদয়, সম্মানপূর্বক বিনীত নিবেদন এই যে...
}
\end{tcolorbox}
\textit{\small (English: coherent, on-topic, correctly structured formal application.)}
\\
\bottomrule
\end{tabular}
\caption{Example of generation incoherence in a compact zero-shot LLM (LLaMA-3.2-3B) versus the corrected output after BLADE fine-tuning, for the prompt "Application for advance leave."}
\label{tab:failure_modes}
\end{table}

The root causes are twofold. First, compact low-resource LLMs, often the only computationally feasible option for low-resource-language deployment, are prone to producing incoherent, off-topic, or degenerate output on structured, format-sensitive tasks without targeted supervision. Second, even fluent output frequently lacks \textit{register-aware supervision}: Bangla encodes social relationships directly into its grammar through a two-tier honorific system, the formal {\fontspec{kalpurush.ttf}আপনি} (Apni: You) obligatory in institutional writing, versus the informal {\fontspec{kalpurush.ttf}তুমি} (Tumi: You) reserved for peers and juniors \ref{tab:spanning_comparison}. Mixing these registers signals a fundamental breakdown in social competence and renders a document unusable in any formal context \citep{snigdha2022representation}. Table~\ref{tab:failure_modes} illustrates the former: a compact zero-shot model produces text entirely unrelated to the prompt rather than a merely mis-registered response.

To address these limitations, we introduce \textbf{BLADE (BangLa Applications and DialoguEs)}, a specialized resource serving a dual role: (i) an instruction-tuning dataset of 4,196 high-quality, expert-annotated prompt-response pairs spanning 2,008 unique topics, and (ii) a rigorous evaluative benchmark for measuring model adherence to Bangla generation reliability, structural, and cultural norms across both large multilingual models and compact low-resource LLMs. BLADE was constructed through a three-tier acquisition pipeline: government-approved textbooks establish canonical formatting; verified web portals provide real-world usage diversity; and author-synthesized examples, cross-validated by native linguistic experts, address complex generation and honorific alignment scenarios underrepresented in static sources. \\ 
\textbf{Our contributions are as follows:}
(1) We introduce \textsc{BLADE}, a publicly available instruction-tuning dataset of 4,196 expert-annotated pairs spanning 2,008 unique topics, built through a three-tier pipeline with native linguistic expert validation, covering educational, professional, and conversational Bangla generation.(2) We benchmark generation reliability and pragmatic gaps in five state-of-the-art multilingual LLMs and four compact low-resource LLMs under zero-shot conditions, systematically documenting coherence failures, register inconsistencies, and structure violations across all tested architectures.(3) We demonstrate that targeted SFT on \textsc{BLADE} yields consistent gains across automatic metrics (BLEU 17.73; chrF 46+), human expert evaluation ($>$4.6/5 on all dimensions), and LLM-as-judge scoring (8.30/10), establishing that cultural data specificity outweighs model scale for reliable, low-resource pragmatic generation.

\section{Related Work}
\label{sec:related_work}

\paragraph{Multilingual LLMs and the Low-Resource Gap.}
Progress in LLMs has been driven by web-scale corpora and instruction datasets such as Common Crawl, \textbf{Alpaca}, and \textbf{Dolly}, with multilingual coverage expanded by projects like HPLT \citep{de-gibert-etal-2024-new}. Despite these gains, a well-documented 25--35\% instruction-following accuracy gap persists between high- and low-resource languages \citep{zeng-etal-2025-marco}, and LLMs consistently fail to outperform classical baselines on extremely low-resource languages in zero-shot settings \citep{cahyawijaya-etal-2024-llms}. Critically, machine-translated instruction data systematically underestimates model capability compared to natively localized equivalents \citep{bawden-yvon-2023-investigating}, motivating our choice to construct BLADE from human-curated, institutionally grounded Bangla sources. \\
\textbf{Honorific Systems and Pragmatic Competence.}
A deeper multilingual failure concerns pragmatic competence: generating text that is socially and institutionally appropriate, not merely grammatical. For languages that grammaticalize social relationships into morphology, register consistency is a necessary condition for functional usability. Japanese LLMs fail to generalize honorific patterns across novel syntactic structures \citep{sekizawa-yanaka-2023-analyzing}, and even frontier models including GPT-4o underperform on high-politeness Javanese registers underrepresented in web-scale data \citep{farhansyah-etal-2025-language}. Arabic evaluation reveals the same failure: surface fluency does not imply cultural coherence \citep{attia-etal-2026-beyond}.

\begin{table*}[t]
\centering
\small
\renewcommand{\arraystretch}{1.5}
\begin{tabular}{m{0.14\textwidth} m{0.40\textwidth} m{0.40\textwidth}}
\specialrule{1.5pt}{0pt}{0pt}
\rowcolor[HTML]{2E3B4E} 
\textbf{\color{white}Topic} & \textbf{\color{white}Content (Bangla)} & \textbf{\color{white}English Translation} \\ 
\specialrule{1pt}{0pt}{0pt}

\centering {\fontspec{kalpurush.ttf}অসুস্থতার কারণে ছুটির আবেদন}\\[4pt] \textit{\footnotesize(Application for sick leave)}
&
\begin{tcolorbox}[colback=blue!5, colframe=blue!20, arc=5pt, boxrule=0.5pt, left=8pt, right=8pt, top=8pt, bottom=8pt]
{\fontspec{kalpurush.ttf}
তারিখঃ ১৮/১০/২০২৪ খ্রিঃ \newline
বরাবর \newline
প্রধান শিক্ষক \newline
ছোলমাইদ হাই স্কুল এন্ড কলেজ \newline
ভাটারা, ঢাকা-১২১২ \newline
\textbf{বিষয়ঃ} বিদ্যালয়ে অনুপস্থিতির জন্য প্রধান শিক্ষকের নিকট আবেদন। \newline
স্যার ......
}
\end{tcolorbox}
&
\begin{tcolorbox}[colback=gray!5, colframe=gray!10, arc=5pt, boxrule=0.5pt, left=8pt, right=8pt, top=8pt, bottom=8pt]
\textit{
Date: 18/10/2024 AD \newline
To \newline
Headmaster \newline
Solmaid High School and College \newline
Vatara, Dhaka-1212 \newline
\textbf{Subject:} Application to the Headmaster regarding absence from school. \newline
Sir......
}
\end{tcolorbox}
\\
\specialrule{1.5pt}{0pt}{0pt}
\end{tabular}
\caption{Example BLADE entry with English translation, illustrating canonical document structure.}
\label{tab:dataset-overview}
\end{table*} 
These findings establish that Bangla's honorific mismatch problem, discussed above, is a cross-linguistically shared structural failure, not a Bangla-specific anomaly. \\
\textbf{Bangla NLP Resources.}
Bangla NLP has progressed from encoder models such as \textit{BanglaBERT} \citep{bhattacharjee-etal-2022-banglabert} to native generative models including \textit{TigerLLM} \citep{raihan-zampieri-2025-tigerllm} and \textit{TituLLMs} \citep{nahin-etal-2025-titullms}, alongside task-specific datasets for NER \citep{paul2025ancholik, mhaske-etal-2023-naamapadam}, captioning \citep{RAHMAN2019636}, and sentiment analysis \citep{islam-etal-2021-sentnob-dataset}. However, all existing resources target classification and extraction; none provides supervision for long-form, format-sensitive, register-consistent generation the precise gap BLADE addresses.

\paragraph{Data Quality over Scale.}
Our core empirical finding that a compact model fine-tuned on BLADE outperforms zero-shot frontier models is grounded in a broader principle: \citet{zhou-etal-2023-lima} show that 1,000 carefully curated examples suffice to match models trained with extensive RLHF, establishing that data quality outweighs quantity for alignment. This has been validated in low-resource multilingual settings, where small-scale high-quality SFT data outperforms larger automatically constructed corpora at the morphological and structural level \citep{iyer-etal-2024-quality}. BLADE provides concrete evidence of this principle for register-sensitive, format-constrained Bangla generation.
\section{BLADE Dataset}

\subsection{Dataset Overview}

The BLADE dataset comprises 4,196 high-quality prompt-response pairs 
constructed specifically for structured Bangla generation, with an average 
response length exceeding 1,300 tokens. The dataset spans 2,008 unique 
topics across two primary task types \textit{application writing} and 
\textit{dialogue generation} covering educational, professional, 
administrative, and conversational domains Table~\ref{tab:topic_distribution}.
BLADE was designed around three core principles absent from existing 
Bangla corpora: (i) \textbf{register awareness} every example enforces 
consistent honorific alignment between salutations, pronouns, verbal 
morphology, and closings; (ii) \textbf{structural fidelity} application 
examples strictly follow canonical Bangla institutional document order 
(Date $\rightarrow$ Addressee $\rightarrow$ Subject $\rightarrow$ 
Salutation $\rightarrow$ Body $\rightarrow$ Closing $\rightarrow$ 
Signature); and (iii) \textbf{topical diversity} topics represent 
real-world Bangla institutional writing encountered by students, 
professionals, and citizens in Bangladesh. An illustrative entry is 
provided in Table~\ref{tab:dataset-overview}.
\begin{table}[t]
\centering
\small
\definecolor{headerblue}{RGB}{210, 226, 242}
\definecolor{lightblue}{RGB}{219, 234, 254}
\definecolor{lighterblue}{RGB}{239, 246, 255}
\definecolor{accentteal}{RGB}{232, 240, 254}
\resizebox{\linewidth}{!}{%
\begin{tabular}{@{}lrlr@{}}
\toprule
\rowcolor{headerblue}
\textcolor{black}{{Applications}} & 
\textcolor{black}{{Topics}} & 
\textcolor{black}{{Dialogues}} & 
\textcolor{black}{{Topics}} \\ 
\midrule
\rowcolor{lightblue}
General    & 675 & General & 476 \\
\rowcolor{lighterblue}
Educational  & 214 & Informal & 170 \\
\rowcolor{lightblue}
Administrative & 169 & Formal  & 141 \\
\rowcolor{lighterblue}
Professional  & 163 &     &   \\ 
\midrule
\rowcolor{accentteal}
\multicolumn{3}{l}{\textcolor{black}{{Total Topics}}} & 
\textcolor{black}{\textbf{2,008}} \\ 
\bottomrule
\end{tabular}%
}
\caption{Distribution of unique topics across domains.}
\label{tab:topic_distribution}
\end{table}

\subsection{Topic Selection}

Topic inclusion followed four criteria: (i) \textit{real-world prevalence} 
in Bangladeshi educational and professional contexts, validated against 
source frequency across textbooks and web portals; (ii) \textit{register 
sensitivity} tasks where honorific errors render output institutionally 
unusable; (iii) \textit{structural complexity}s multi-component documents 
beyond a single paragraph; and (iv) \textit{under representation} in 
existing pretraining corpora. Topics already well-covered by generic 
web-scale data (e.g., news, Wikipedia) were explicitly deprioritized. 
The resulting distribution spans 1,221 application topics and 787 
dialogue topics Table~\ref{tab:topic_distribution}.

\subsection{Data Collection}
\label{sec:data-collection}

Collection followed a three-tier acquisition strategy, each tier governed by distinct sourcing and verification protocols. Full annotation guidelines, annotator profiles, inter-annotator agreement statistics, and quality control criteria are provided in Appendix~\ref{appendix:annotation}.

\paragraph{Tier 1: Institutional Textbooks (1,972 samples).}
Canonical application structures were extracted from nine 
government-approved NCTB secondary and higher secondary textbooks, 
establishing a gold standard for institutional formatting. Each entry 
was verified by two independent annotators for structural completeness 
(all seven mandatory document components present in canonical order) 
before inclusion.

\paragraph{Tier 2: Public Web Portals (1,382 samples).}
Real-world examples were curated from 14 verified Bangla educational 
web portals Table~\ref{tab:website_sources_url_only}. Every entry 
underwent manual verification against structural completeness and 
honorific consistency criteria. Approximately 23\% of candidate web 
entries were discarded during this process due to register 
inconsistencies or malformed structures.

\paragraph{Tier 3: Author-Synthesized Examples (842 samples).}
To address complex scenarios underrepresented in static sources particularly multi-turn dialogues requiring sustained honorific consistency 842 examples were purpose-built by the authors and cross-validated by two external native linguistic experts specializing in Bangla philology. Each Tier 3 example passed a two-stage review: 
author-annotator drafting followed by independent expert adjudication.

\subsection{Data Preprocessing}
\begin{figure*}[t]
\centering
\includegraphics[width=\textwidth]{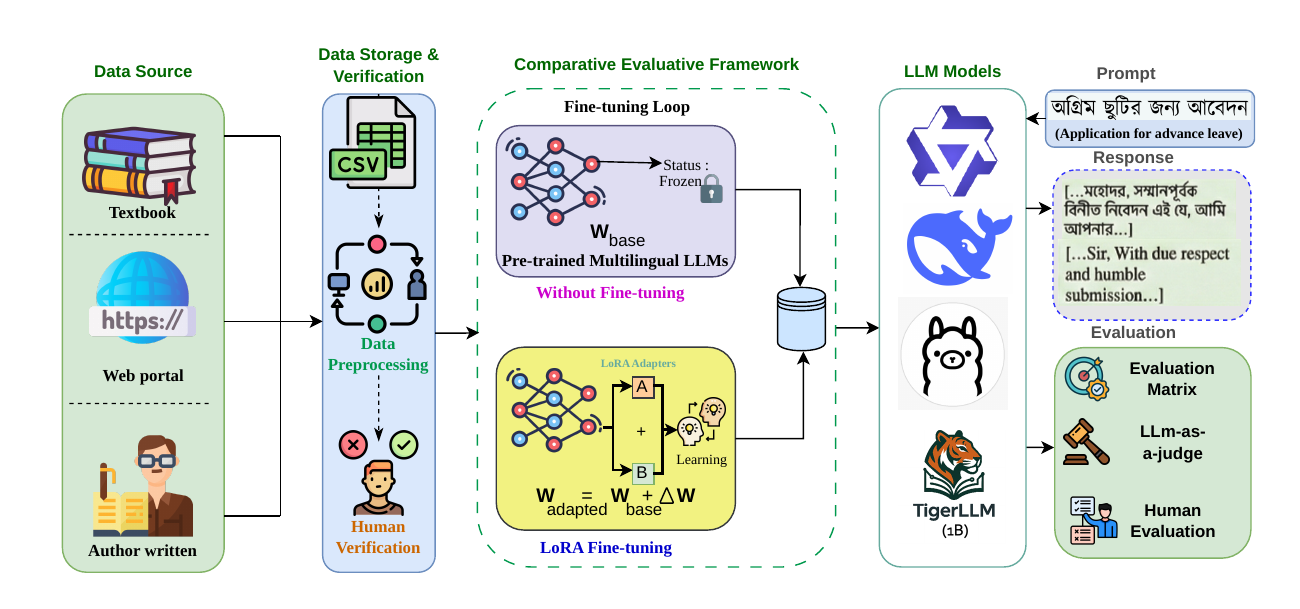}
\caption{BLADE methodology: three-tier data collection, LoRA-based fine-tuning of pre-trained multilingual LLMs, and a three-way comparative evaluation framework}
\label{tab:dataset-sample}
\end{figure*}

Raw entries are cleaned using regular expressions preserving Bangla Unicode (U+0980–U+09FF), reformatted into the instruction-tuning template described in Section~\ref{sec:fine-tune}, and partitioned into a 80\% training set 3,356 examples 10\% testing set 420 examples and 10\% validation set 420 examples via stratified sampling across domain categories with a fixed random seed. The complete pipeline is illustrated in 
Figure~\ref{tab:dataset-sample}.

\section{Methodology}
Here we describe the experimental setups for both zero-shot evaluation and supervised fine-tuning.
\subsection{Zero-Shot Setup}
\label{sec:zero-shot}
To establish a baseline for multilingual LLMs, we conducted a zero-shot evaluation on our dataset, The dataset coverage all domain categories Table~\ref{tab:topic_distribution}. Each model was queried using the identical structured template described in Section~\ref{prompt_setting} to isolate true instruction-following capability without task-specific demonstrations. It is important to note that zero-shot models are evaluated on a prompt format they have not been explicitly trained on; consequently, some portion of the observed performance gap may reflect \textit{format unfamiliarity} rather than Bangla specific incapability alone.

For inference, we utilized the Groq 
API\footnote{\url{https://console.groq.com/}} to evaluate \texttt{Llama-4-Scout-17B}, \texttt{Gemma2-9B}, and \texttt{Kimi-K2-32B}. Google AI Studio was used to evaluate \texttt{Gemini-2.5-Flash} and \texttt{Gemini-2.0-Flash}. Performance was benchmarked against BLADE ground-truth responses.
\subsection{Fine-tuning Setup}
\label{sec:fine-tune}
To adapt pre-trained LLMs for Bangla instruction following, we employed Parameter-Efficient Fine-Tuning (PEFT) using 4-bit NormalFloat (NF4) quantization and Low-Rank Adaptation (LoRA) \citep{hu2021lora}. We selected four architectures \texttt{DeepSeek-R1-Distill-Llama-8B}, \texttt{Qwen2-1.5B}, \texttt{Llama-3.2-3B-Instruct}, and \texttt{TigerLLM-1B-it} to evaluate BLADE across varying scales and training paradigms. \texttt{TigerLLM} serves as a Bangla-native baseline; the remaining models represent state-of-the-art multilingual architectures at the 1B--8B scale, chosen to test efficacy under resource-constrained deployment conditions.

All models were loaded in 4-bit precision with FP16 computation via the \texttt{Unsloth} library on dual NVIDIA Tesla T4 GPUs (32~GB total VRAM). SFT was conducted for 2 epochs with a 2048 token sequence length using \texttt{SFTTrainer}, batch size 2, cosine learning rate decay with 5 warm-up steps at peak $2{\times}10^{-5}$, and AdamW optimizer. LoRA was configured with rank $r{=}16$, $\alpha{=}32$, dropout 0.05, applied to both attention and MLP projection layers. The prompt template have (Listing~\ref{prompt_setting}) enforced canonical document order (Subject $\rightarrow$ Salutation $\rightarrow$ Body $\rightarrow$ Closing) and register consistency, with loss computed only over the \texttt{<assistant>} span.
\begin{tcolorbox}[
  title=\textbf{Prompt Template}, 
  colback=teal!5!white,
  colframe=teal!80!black,
  coltitle=white,
  boxrule=1pt, 
  arc=4pt,
  shadow={2mm}{-2mm}{0mm}{black!20},
  label=prompt_setting
]
\small\ttfamily
<system>\ \\
order: Subject $\rightarrow$ Salutation $\rightarrow$ Body $\rightarrow$ Closing. Keep consistent honorifics.

<user>\ \\
\textbf{Task:} Formal application.\\
\textit{Subject}: {\fontspec{kalpurush.ttf} অগ্রিম ছুটির জন্য আবেদন}\\
\textit{Addressee}: {\fontspec{kalpurush.ttf} প্রধান শিক্ষক, এক্সওয়াইজেড উচ্চ বিদ্যালয়}\\
\textit{Applicant}: {\fontspec{kalpurush.ttf} মোঃ রহমান, নবম শ্রেণি, রোল ১২}\\
\textit{Reason}: {\fontspec{kalpurush.ttf} ডাক্তার দেখাতে শহরে যেতে হবে}\\
\textit{Dates}: {\fontspec{kalpurush.ttf} ১০–১২ আষাঢ় ১৪৩২}\\
\textit{Closing}: {\fontspec{kalpurush.ttf} বিনীতভাবে}\\
\textbf{Constraints:} 150--200 words;Bangla digits;nocode-mixing.

<assistant>\ \\
\end{tcolorbox}

\definecolor{tableheader}{RGB}{210, 226, 242}
\definecolor{tablerow}{RGB}{245, 245, 245}
\definecolor{highlight}{RGB}{232, 240, 254}
\definecolor{fmhighlight}{RGB}{200, 218, 245}

\begin{table*}[t]
\centering
\small
\setlength{\tabcolsep}{6pt}
\renewcommand{\arraystretch}{1.2}
\begin{tabular}{llccccc}
\toprule
\rowcolor{tableheader}
\textbf{Model} & \textbf{Condition} & \textbf{BLEU} ($\uparrow$) & \textbf{ROUGE-L} ($\uparrow$) & \textbf{chrF} ($\uparrow$) & \textbf{WER} ($\downarrow$) & \textbf{BERTScore} ($\uparrow$) \\
\midrule

\rowcolor{white}
Kimi-K2-32B & Zero-Shot & 4.55 & 0.670 & 37.15 & 2.283 & 0.722 \\
\rowcolor{fmhighlight}
Kimi-K2-32B & FM & 5.12\textsubscript{\textcolor{teal!70!black}{$\uparrow$12.5\%}} & \textbf{0.731}\textsubscript{\textcolor{teal!70!black}{$\uparrow$9.1\%}} & 40.38\textsubscript{\textcolor{teal!70!black}{$\uparrow$8.7\%}} & 1.974\textsubscript{\textcolor{teal!70!black}{$\downarrow$13.5\%}} & 0.739\textsubscript{\textcolor{teal!70!black}{$\uparrow$2.4\%}} \\
\midrule

\rowcolor{white}
LLaMA-4-Scout-17B & Zero-Shot & 6.30 & 0.290 & 35.92 & \textbf{1.729} & 0.715 \\
\rowcolor{fmhighlight}
LLaMA-4-Scout-17B & FM & 7.43\textsubscript{\textcolor{teal!70!black}{$\uparrow$17.9\%}} & 0.348\textsubscript{\textcolor{teal!70!black}{$\uparrow$20.0\%}} & 39.11\textsubscript{\textcolor{teal!70!black}{$\uparrow$8.9\%}} & 1.521\textsubscript{\textcolor{teal!70!black}{$\downarrow$12.0\%}} & 0.728\textsubscript{\textcolor{teal!70!black}{$\uparrow$1.8\%}} \\
\midrule

\rowcolor{white}
Gemini-2.0-Flash & Zero-Shot & 3.73 & 0.250 & 34.45 & 3.895 & 0.711 \\
\rowcolor{fmhighlight}
Gemini-2.0-Flash & FM & 4.58\textsubscript{\textcolor{teal!70!black}{$\uparrow$22.8\%}} & 0.317\textsubscript{\textcolor{teal!70!black}{$\uparrow$26.8\%}} & 38.27\textsubscript{\textcolor{teal!70!black}{$\uparrow$11.1\%}} & 3.114\textsubscript{\textcolor{teal!70!black}{$\downarrow$20.1\%}} & 0.726\textsubscript{\textcolor{teal!70!black}{$\uparrow$2.1\%}} \\
\midrule

\rowcolor{white}
Gemma2-9B & Zero-Shot & 4.69 & 0.100 & 34.80 & 1.765 & 0.706 \\
\rowcolor{fmhighlight}
Gemma2-9B & FM & 5.47\textsubscript{\textcolor{teal!70!black}{$\uparrow$16.6\%}} & 0.134\textsubscript{\textcolor{teal!70!black}{$\uparrow$34.0\%}} & 37.93\textsubscript{\textcolor{teal!70!black}{$\uparrow$9.0\%}} & 1.483\textsubscript{\textcolor{teal!70!black}{$\downarrow$16.0\%}} & 0.719\textsubscript{\textcolor{teal!70!black}{$\uparrow$1.8\%}} \\
\midrule

\rowcolor{highlight}
Gemini-2.5-Flash & Zero-Shot & \textbf{7.88} & 0.250 & \textbf{41.04} & 1.934 & \textbf{0.728} \\
\rowcolor{fmhighlight}
\textbf{Gemini-2.5-Flash} & \textbf{FM} & \textbf{9.14}\textsubscript{\textcolor{teal!70!black}{$\uparrow$16.0\%}} & 0.390\textsubscript{\textcolor{teal!70!black}{$\uparrow$56.0\%}} & \textbf{45.67}\textsubscript{\textcolor{teal!70!black}{$\uparrow$11.3\%}} & \textbf{1.203}\textsubscript{\textcolor{teal!70!black}{$\downarrow$37.8\%}} & \textbf{0.751}\textsubscript{\textcolor{teal!70!black}{$\uparrow$3.2\%}} \\

\midrule
\rowcolor{tableheader!50}
\multicolumn{7}{c}{\textit{Human Evaluation (1--5 scale: Structure / Fluency / Cultural Alignment)}} \\
\midrule
\rowcolor{white}
Zero-Shot Baseline & & \multicolumn{5}{c}{1.85 / 2.40 / 2.15} \\
\rowcolor{fmhighlight}
FM Baseline & & \multicolumn{5}{c}{4.20 / 3.80 / 3.50} \\
\bottomrule
\end{tabular}
\caption{Zero-shot vs.\ FM(Format-Matched) performance across all models. Subscript percentages indicate relative improvement of FM over Zero-Shot ($\uparrow$ higher is better, $\downarrow$ lower is better for WER). Best automatic metric scores in \textbf{bold}. Human evaluation reports Structure / Fluency / Cultural Alignment (Apni/Tumi register)  respectively.}
\label{tab:zeroshot_final}
\end{table*}

\definecolor{headerblue}{RGB}{232, 235, 245}
\definecolor{rowgray}{RGB}{248, 248, 248}
\definecolor{metricgroup}{RGB}{240, 240, 245}

\begin{table*}[t]
\centering
\small
\setlength{\tabcolsep}{5pt}
\renewcommand{\arraystretch}{1.2} 
\begin{tabular*}{\textwidth}{@{\extracolsep{\fill}} l rr rr rr rr rr rr rr}
\toprule
\rowcolor{headerblue}
& \multicolumn{2}{c}{\textbf{BLEU} $\uparrow$}
& \multicolumn{2}{c}{\textbf{chrF} $\uparrow$}
& \multicolumn{2}{c}{\textbf{WER} $\downarrow$}
& \multicolumn{2}{c}{\textbf{BERTScore} $\uparrow$}
& \multicolumn{2}{c}{\textbf{ROUGE-L} $\uparrow$}
& \multicolumn{2}{c}{\textbf{Human} $\uparrow$}
& \multicolumn{2}{c}{\textbf{Judge} $\uparrow$} \\
\cmidrule(lr){2-3}\cmidrule(lr){4-5}\cmidrule(lr){6-7}
\cmidrule(lr){8-9}\cmidrule(lr){10-11}
\cmidrule(lr){12-13}\cmidrule(lr){14-15}
\rowcolor{headerblue}
\textbf{Model} & Base & SFT & Base & SFT & Base & SFT 
& Base & SFT & Base & SFT & Base & SFT & Base & SFT \\
\midrule

\textbf{DeepSeek-8B}  
& 0.78 & \textbf{17.73} 
& 9.85 & \textbf{45.87} 
& 1.62 & 1.25 
& 0.582 & 0.743 
& 0.164 & \textbf{0.71}
& 2.13 & \textbf{4.74}
& 3.8 & \textbf{8.9} \\

\rowcolor{rowgray}
\textbf{Qwen2-1.5B}   
& 0.86 & 16.28 
& 11.08 & 36.13 
& 1.34 & \textbf{0.84} 
& 0.575 & \textbf{0.748} 
& 0.105 & 0.55
& 2.05 & 4.63
& 3.1 & 8.4 \\

\textbf{LLaMA3.2-3B}  
& 1.96 & 15.30 
& 24.88 & \textbf{46.60} 
& 1.66 & 1.33 
& 0.632 & 0.742 
& 0.101 & 0.62
& 2.21 & 4.70
& 4.2 & 8.7 \\

\rowcolor{rowgray}
\textbf{TigerLLM-1B}  
& 1.20 & 7.15 
& 10.90 & 33.83 
& 1.59 & 1.18 
& 0.240 & 0.716 
& 0.086 & 0.35
& 1.80 & 4.61
& 1.5 & 7.2 \\
\bottomrule
\end{tabular*}
\caption{Evaluation results before (Base) and after (SFT) 
fine-tuning on BLADE. \textbf{Human} scores are averaged 
across Structure, Fluency, and Cultural Alignment (1--5 scale). LLM-as-judge ratings (1--10 scale) 
using GPT-4.1 across 450 samples. Best SFT result 
per metric in \textbf{bold}.}
\label{tab:results_transposed}
\end{table*}

\subsection{Evaluation Metrics}
\label{evaluation_mat}

We evaluate model outputs using five complementary metrics that 
collectively capture lexical, structural, and semantic similarity 
between generated and reference texts. \textbf{BLEU} 
\citep{papineni-etal-2002-bleu} measures $n$-gram overlap with 
a brevity penalty. \textbf{chrF} \citep{popovic-2015-chrf} computes 
a character $n$-gram F-score, making it particularly robust for 
morphologically rich languages like Bangla. \textbf{ROUGE-L} 
\citep{lin-2004-rouge} measures the longest common subsequence 
between reference and hypothesis. \textbf{WER} 
\citep{ali-renals-2018-word} measures normalized edit distance 
between hypothesis and reference, where lower scores indicate 
fewer structural errors. \textbf{BERTScore} 
\citep{zhang2019bertscore} evaluates semantic similarity using 
contextual embeddings from a pretrained model.
Given the known limitations of similarity-based metrics in 
capturing semantic coherence and cultural alignment, automatic 
metric results are complemented by human expert evaluation and 
LLM-as-judge scoring, both reported alongside automatic metrics 
in Section~\ref{sec:results}.

\section{Result Analysis}\label{sec:results}
We benchmark our dataset through two assessments: (1) evaluating state-of-the-art models in a zero-shot setting and (2) comparing model performance before(base model) and after Supervised Fine-Tuning (SFT). This dual approach highlights the dataset’s utility and the knowledge gap it addresses in NLP.

\subsection{Zero-Shot and Format-Matched Results}
Table~\ref{tab:zeroshot_final} presents performance across five state-of-the-art
architectures under zero-shot and format-matched prompting. Zero-shot results confirm
a significant generation and pragmatic gap: no model exceeds 8 BLEU, with three
recurring failure modes: \textit{generation incoherence} (off-topic or degenerate
output, as illustrated in Table~\ref{tab:failure_modes}), \textit{structural
displacement} (misplaced date, addressee, or closing blocks), and \textit{honorific
mismatch} (inconsistent \textit{Apni/Tumi} register within a single document).

\paragraph{Zero-Shot Performance.}
\textit{Gemini-2.5-Flash} achieves the strongest zero-shot performance (BLEU 7.88,
chrF 41.04, BERTScore 0.728), yet confirms that even the strongest closed-source
baseline fails to produce functionally usable Bangla documents without targeted
supervision. \textit{Kimi-K2-32B} leads on ROUGE-L (0.67), though its lower BLEU
reveals weak local $n$-gram precision. \textit{LLaMA-4-Scout-17B} achieves the lowest
WER (1.73), but this does not translate to overall generation quality, confirming that
word-level accuracy alone is insufficient for structured Bangla generation.

\paragraph{Format-Matched Prompting.}
Explicit structural constraints yield consistent improvements across all five models.
\textit{Gemini-2.5-Flash} continues to lead: BLEU 7.88 $\rightarrow$ 9.14
($\uparrow$16.0\%), chrF 41.04 $\rightarrow$ 45.67 ($\uparrow$11.3\%), BERTScore
0.728 $\rightarrow$ 0.751 ($\uparrow$3.2\%), and WER 1.934 $\rightarrow$ 1.203
($\downarrow$37.8\%)   the largest absolute error reduction across all models.
\textit{Gemini-2.0-Flash} shows the most pronounced BLEU gain ($\uparrow$22.8\%),
while \textit{Gemma2-9B} records the largest ROUGE-L gain ($\uparrow$34.0\%),
reflecting improved long-range structural recall. \textit{LLaMA-4-Scout-17B} maintains
its WER advantage (1.521), confirming stable word-level alignment independent of
prompting strategy. BERTScore gains remain modest (1.8\%--3.2\%), indicating semantic fidelity is largely
preserved in zero-shot outputs and that format-matching primarily addresses surface
structural and $n$-gram precision failures. These results confirm that explicit
structural scaffolding is a lightweight yet effective intervention for closing the
pragmatic gap in formal Bangla document generation, without requiring parameter updates
or fine-tuning.
\subsection{Fine-tuning Results}

Table~\ref{tab:results_transposed} presents automatic metrics, human evaluation, and LLM-as-judge scores before and after SFT across all four architectures. The consistent gains across every model and every evaluation dimension confirm that BLADE provides 
a high-fidelity, transferable training signal for low-resource structured generation.\textbf{ DeepSeek-8B} exhibits the most pronounced improvement: a 22-fold BLEU increase (0.78 $\rightarrow$ 17.73), chrF rising 
from 9.85 to 45.87, and LLM-judge score jumping from 3.8 to 8.9. This delta reflects a fundamental shift in the model's ability to handle Bangla's inflectional morphology, structural syntax, and register consistency simultaneously.\textbf{Qwen2-1.5B} achieves the lowest post-SFT WER (0.84) and highest BERTScore (0.748), indicating superior semantic alignment despite its compact size. Its performance-to-parameter ratio establishes it as the strongest candidate for resource-constrained deployment. \textbf{LLaMA3.2-3B} attains the peak post-SFT chrF (46.60), confirming that targeted instruction-tuning on BLADE allows smaller models to rival larger multilingual baselines on character-level structural accuracy.\textbf{TigerLLM-1B-it}, despite starting from the weakest baseline (1.20 BLEU, judge score 1.5), achieves a 3$\times$ chrF increase post-SFT and a judge score of 7.2 validating that BLADE's expert-curated supervision is effective even for Bangla-native architectures that already possess foundational language knowledge.
Across all models, the average LLM-judge score rises from 3.15 to 8.30, and human evaluation scores exceed 4.6/5 on all three dimensions post-SFT. The strong alignment between automatic metrics, human judgment, and LLM-judge scoring provides convergent validity for BLADE's effectiveness as both a training resource and evaluation benchmark.

\subsubsection{Ablation Study}

We investigate the drivers of these gains by fine-tuning the base checkpoints on the BLADE training split. Metrics are averaged across models. Our final configuration (Table~\ref{tab:ablation_summary_compact}) uses: AdamW (LR $2{\times}10^{-5}$, cosine decay, 3\% warmup), batch size 2 (gradient accumulation), sequence length 2048 with packing, LoRA $r{=}16$ ($\alpha{=}32$) on attention/MLP, label smoothing 0.1, and a format-aware template with role tags.

Training on larger fractions of BLADE (25\%--100\%) yields monotonic gains, with the steepest jump at 50\% (avg.\ $\Delta$BLEU {+}5.4; $\Delta$chrF {+}9.1). The full set delivers the lowest WER, underscoring that targeted supervision drives structural fidelity. Prompt design is equally critical: a format-aware template surfacing fields like \texttt{Subject} substantially reduces WER ($-0.18$) and boosts ROUGE-L ({+}0.08), while explicit role tags further improve BERTScore and chrF ({+}2.7).
Context length is decisive; truncating to 512 tokens degrades long-form fidelity, and 1024 tokens often miss signatures given the average response length of 1,300. Extending to 2048 with sliding-window packing ensures complete structure, yielding the strongest structural scores (avg.\ $\Delta$chrF {+}3.5). Adapter capacity also matters: LoRA $r{=}16$ outperforms $r{=}8$ (avg.\ $\Delta$BLEU {+}1.3) without the overfitting seen at $r{=}32$. Cosine decay with a peak LR of $2{\times}10^{-5}$ proved robust across architectures, whereas higher rates degraded stability. Label smoothing (0.1) improves semantic alignment (avg.\ $\Delta$BS {+}0.006) and mitigates brittle copying. Applying LoRA to both attention \emph{and} MLP projections beats attention-only adaptation (avg.\ $\Delta$BLEU {+}0.9; $\Delta$chrF {+}1.8). Finally, mixed precision (bf16/fp16) matches full-precision quality while enabling larger batches and slightly improving WER ($-0.03$) via stable optimization.

\begin{table}[t]
\centering
\setlength{\tabcolsep}{3pt}
\small
\begin{tabular}{lccc}
\toprule
\textbf{Ablation (avg.)} & \textbf{$\Delta$BL} & \textbf{$\Delta$cF} & \textbf{$\Delta$WER} \\
\midrule
+ FmtTpl (vs.\ min)     & {+}1.1 & {+}2.4 & $-$0.12 \\
+ Roles           & {+}0.6 & {+}2.7 & $-$0.06 \\
Ctx~2048{+}Pack (vs.\ 512)  & {+}1.4 & {+}3.5 & $-$0.09 \\ 
LoRA $r{=}16$ (vs.\ $r{=}8$) & {+}1.3 & {+}1.8 & $-$0.07 \\
Cosine LR @ $2{\times}10^{-5}$ & {+}0.8 & {+}1.2 & $-$0.05 \\
LS $=0.1$ (vs.\ 0.0)     & {+}0.5 & {+}0.9 & $-$0.03 \\
LoRA Attn{+}MLP (vs.\ Attn) & {+}0.9 & {+}1.8 & $-$0.05 \\
\bottomrule
\end{tabular}
\caption{Ablation summary table. Deltas are averaged across models and computed relative to a vanilla SFT baseline (same data, minimal prompt, max len 512, LoRA $r{=}8$, linear LR, no label smoothing). \textbf{Abbrev.:} BL{=}BLEU, cF{=}chrF, WER{=}Word Error Rate, \textit{FmtTpl}{=}format-aware template, \textit{Roles}{=}role tags, \textit{Ctx}{=}context length, \textit{Pack}{=}sliding-window packing, \textit{LR}{=}learning rate, \textit{LS}{=}label smoothing, \textit{Attn}{=}attention, \textit{MLP}{=}feed-forward block. Positive $\Delta$ means improvement; lower WER is better.}
\label{tab:ablation_summary_compact}
\end{table}

\subsection{LLM-as-Judge Procedure}
\label{sec:llm_judge}
To assess functional usability beyond automatic metrics, we employed \texttt{GPT-4.1}
as an LLM judge across 450 randomized samples. The judge was provided with the input
prompt (Appendix~\ref{appendix:judge}), the rationale for selecting \texttt{GPT-4.1}
(Appendix~\ref{gemini-judge}), a reference response, and the model output, and
instructed to score on a 1--10 scale against three criteria: (i) \textit{structural correctness}   presence and ordering of all mandatory document components; (ii) \textit{register consistency} honorific alignment throughout; and (iii) \textit{semantic relevance}   whether the output addresses the stated prompt purpose. The judge prompt was fixed across all 450 samples to ensure scoring consistency, and outputs were parsed programmatically; malformed scores were re-queried once before exclusion.
\subsection{Human Evaluation}
\label{sec:human_eval}
Three native Bangla-speaking philology experts conducted a double-blind assessment of
100 stratified samples, rating outputs on a 5-point Likert scale across: (i)
\textit{Structural Integrity}   adherence to canonical document formatting; (ii)
\textit{Fluency}   grammatical correctness and naturalness; and (iii)
\textit{Cultural Alignment}   honorific register consistency. Inter-annotator
agreement was robust (Spearman's $\rho{=}0.84$, Kendall's $\tau{=}0.76$), and
post-SFT scores exceeded 4.6/5 across all dimensions
(Table~\ref{tab:results_transposed}), confirming that BLADE fine-tuning enables
models to produce institutionally appropriate, functionally usable Bangla documents.

\section{Discussion}
\label{sec:discussion}
The central finding of BLADE is straightforward: for low-resource languages like Bangla,
cultural specificity of instruction-tuning data matters more than model scale.
\texttt{Qwen2-1.5B} fine-tuned on BLADE achieves BERTScore 0.748 and WER 0.84,
outperforming zero-shot \texttt{Gemini-2.5-Flash} on both metrics despite being orders
of magnitude smaller, confirming that expert-curated, register-aware supervision
unlocks capabilities that general-purpose pretraining at any scale cannot substitute.

\paragraph{Qualitative Generation and Pragmatic Competence Shift}
Zero-shot outputs fail in three institutionally fatal ways: \textit{generation
incoherence} (off-topic or degenerate output, as illustrated in
Table~\ref{tab:failure_modes}), \textit{structural displacement} (missing or misplaced
date blocks, subject lines, or closings), and \textit{honorific mismatch}, where models
oscillate between formal {\fontspec{kalpurush.ttf}আপনি (Apni)} and informal
{\fontspec{kalpurush.ttf}তুমি (Tumi)} registers within a single document. No institution
would accept such output; the strongest zero-shot model scores only 1.85/5 on
Structural Integrity. Post-SFT, all three failure modes are eliminated: models produce
coherent, on-topic responses with correctly formatted date and addressee blocks,
maintain formal register across all verb conjugations, and close with culturally
appropriate markers such as {\fontspec{kalpurush.ttf}বিনীতভাবে জানাচ্ছি (I respectfully
inform you)}, representing a qualitative shift from outputs that \textit{appear} fluent
to outputs that are \textit{functionally usable}. chrF captures this most faithfully:
sensitive to the inflectional suffixes and agglutinative morphemes carrying honorific
information in Bangla, a jump from 9.85 to 45.87 for \texttt{DeepSeek-8B} reflects
correct conjugations, consistent honorific suffixes, and structurally complete
documents, explaining its strongest correlation with human expert scores.

\paragraph{Validation Across Evaluation Dimensions}
Three independent evaluation channels, human philologists operating double-blind, an
LLM judge, and reference-based automatic metrics, converge on the same conclusion.
Human evaluators awarded post-SFT outputs above 4.6/5 across all dimensions, with
Structural Integrity showing the largest gain (1.85 $\rightarrow$ 4.72). The LLM-judge
average rose from 3.15 to 8.30 across 450 samples, tracking quantitative deltas
precisely: \texttt{DeepSeek-8B}'s 22-fold BLEU increase corresponds to the highest
judge score (8.9), \texttt{LLaMA3.2-3B}'s peak chrF of 46.60 to a score of 8.7, and
\texttt{TigerLLM-1B}'s modest gains to the lowest post-SFT score (7.2). This
convergence rules out metric-specific artifacts and confirms that BLADE produces
genuine improvement in functional usability, the exact quality that zero-shot
multilingual pretraining systematically fails to provide.
\section{Conclusion}
We presented \textbf{BLADE}, a 4,196-pair instruction-tuning dataset encoding the
structural, register, and cultural conventions for usable Bangla generation. While
baseline multilingual models fell short, supervised fine-tuning on BLADE delivered
large, consistent gains across architectures \textit{including both multilingual and
Bangla-native models}, converting surface-level fluency into outputs meeting real-world
expectations. Our findings confirm that carefully curated, domain-specific supervision
unlocks capabilities that generic pretraining and parameter count cannot, enabling
compact models to rival larger ones on practical tasks. We aim for BLADE to catalyze
structure-aware, register-sensitive generation in Bangla and other underserved
languages, encouraging evaluations that pair automatic metrics with format checks and
human judgments, with future extensions to additional genres and integrated validators.

\newpage
\clearpage
\section*{Limitations}
\label{sec:limitations}

Our study has several constraints that present 
opportunities for future refinement. \\ 
First, all models were fine-tuned for only two 
epochs due to hardware constraints (dual NVIDIA 
Tesla T4 GPUs, 32 GB total VRAM), which may 
have limited convergence and overall model 
stability. These same resource constraints 
precluded exhaustive hyperparameter search and 
more comprehensive ablation studies across a 
wider range of configurations.
Second, the BLADE dataset is currently biased 
toward formal Bangla registers, as informal and 
dialectal sources were less accessible during 
collection. This limits the generalizability of 
fine-tuned models to informal conversational 
contexts.

Future work will address these limitations by 
extending training duration with more capable 
hardware, expanding dataset coverage to include 
informal Bangla dialects and regional varieties, 
and conducting broader evaluations across 
additional document genres and task types.

\bibliography{custom}

\clearpage
\newpage
\appendix

\section{Annotation Protocol and Quality Control}
\label{appendix:annotation}

\subsection{Annotator Profile}

The dataset was constructed and validated by seven annotators: five 
paper authors with backgrounds in computer science and Bangla 
linguistics, and two external native linguistic experts specializing 
in Bangla philology and sociolinguistics. All annotators are native 
Bangla speakers with formal education conducted in Bangla. External 
experts were compensated at a standard research rate, informed of 
the research purpose prior to participation, and provided explicit 
written consent for inclusion.

\subsection{Annotation Guidelines}

Annotators followed a written guideline document evaluating each 
entry across three dimensions:

\paragraph{1. Structural Completeness.}
Structural completeness is a binary pass/fail criterion evaluated 
against a document-type-specific checklist. A response is 
structurally complete if and only if it contains \textit{all} 
mandatory components for its document type in the correct canonical 
order. For formal applications, the seven mandatory components are:

\begin{enumerate}
  \item Date in Bangla format 
  (e.g., {\fontspec{kalpurush.ttf}১৮/১০/২০২৪ খ্রিঃ})
  \item Addressee block: recipient name/title, institution, 
  and address
  \item Subject line 
  ({\fontspec{kalpurush.ttf}বিষয়ঃ})
  \item Formal salutation 
  ({\fontspec{kalpurush.ttf}মহোদয়} or equivalent)
  \item Body: minimum two paragraphs — context statement 
  and formal request
  \item Formal closing 
  ({\fontspec{kalpurush.ttf}বিনীত} or equivalent)
  \item Applicant signature block: name, class/position, 
  roll/ID, institution
\end{enumerate}

A response missing any component, or presenting components out 
of canonical order, fails this criterion and is either corrected 
(minor errors, e.g., missing date field) or discarded 
(structural malformation or register inconsistency).

\paragraph{2. Honorific Consistency.}
A response passes honorific consistency if all second-person 
pronouns, associated verb forms, and relational terms maintain 
a single register throughout. Formal register requires exclusive 
use of {\fontspec{kalpurush.ttf}আপনি (Apni: you)} and its associated 
verb conjugations. Any occurrence of informal forms 
{\fontspec{kalpurush.ttf}তুমি (Tumi: you)} or 
{\fontspec{kalpurush.ttf}তুই (Tui: you)} within a formally-labeled 
entry constitutes an honorific violation and triggers rejection 
or correction. For dialogue entries, the required register 
(formal/informal) is determined by the topic label and must 
remain consistent throughout all turns.

\paragraph{3. Cultural and Contextual Accuracy.}
Content must reflect realistic Bangladeshi institutional 
contexts: plausible institution names, dates in correct 
Bangla calendar or AD format, and discourse markers 
appropriate to the document type 
(e.g., {\fontspec{kalpurush.ttf}অতএব} for formal petition 
closings, {\fontspec{kalpurush.ttf}বিনীতভাবে জানাচ্ছি} 
for formal body openings).

\subsection{Annotation Workflow}

\paragraph{Tier 1 and Tier 2 entries} underwent two-annotator 
verification against all three criteria. A stratified random 
sample of 15\% of these entries 630 examples was independently 
re-verified by a second annotator to estimate verification 
reliability.

\paragraph{Tier 3 entries} followed a stricter two-stage workflow: 
(i) draft production by one author-annotator, and (ii) independent 
review by a second annotator against all three criteria. 
Disagreements were escalated to one of the two external linguistic 
experts for final adjudication. No Tier 3 entry was included 
without passing both stages.

\subsection{Inter-Annotator Agreement}

Cross-validation on the 630-entry random sample yielded 
Cohen's $\kappa = 0.81$ for structural completeness judgments 
and $\kappa = 0.79$ for honorific consistency judgments, 
indicating strong inter-annotator agreement \citep{landis1977measurement}. 
Disagreements on the cross-validation sample were resolved 
by majority vote among three annotators.

\subsection{Quality Control Mechanisms}

Three dataset-level quality control mechanisms were applied 
following per-entry annotation:

\paragraph{Automated Register Audit.}
All 4,196 entries underwent automated pre-screening using 
regular expression filters to flag occurrences of informal 
pronoun forms {\fontspec{kalpurush.ttf}তুমি, তুই, তোমার, তোর} means you, within formally-labeled entries. Flagged entries were 
manually reviewed; confirmed violations were corrected or 
discarded.

\paragraph{Structural Completeness Audit.}
A secondary pass verified that all seven mandatory components 
were present for every formal application entry, using the 
checklist defined in Section~\ref{appendix:annotation}.2. 
Entries failing this audit after the initial annotation 
phase were returned to annotators for correction.

\paragraph{Semantic Diversity Check.}
To prevent topic redundancy across 2,008 topics, character 
n-gram similarity was computed across all topic labels. 
Topic pairs with similarity exceeding 0.85 were flagged 
for manual review; confirmed near-duplicates were merged 
or one was discarded, ensuring each topic represents a 
genuinely distinct writing scenario.

\section{More Details on LLM-as-Judge}
\label{appendix:judge}

\subsection{Prompt Template}

\begin{tcolorbox}[
  title=\textbf{LLM-as-Judge Evaluation Prompt},
  colback=blue!5!white,
  colframe=blue!40!black,
  coltitle=white,
  boxrule=1pt,
  arc=4pt
]
\small\ttfamily
You are evaluating a Bangla text generation model.\\[4pt]
\textbf{Input Prompt:} \{prompt\}\\
\textbf{Reference Response:} \{reference\}\\
\textbf{Model Output:} \{output\}\\[4pt]
Score the model output from 1--10 based on:\\
(i) \textbf{Structural Correctness}: Are all mandatory 
document components present in canonical order?\\
(ii) \textbf{Register Consistency}: Are honorifics 
(Apni/Tumi:you) used consistently throughout?\\
(iii) \textbf{Semantic Relevance}: Does the output 
address the stated prompt purpose?\\[4pt]
Return only a single integer score between 1 and 10.
\end{tcolorbox}
\subsection{Rationale for Selecting GPT-4.1 as LLM Judge}
\label{gemini-judge}
AS our zero-shot evaluation included two Gemini-family models, Gemini-2.0-Flash and Gemini-2.5-Flash as evaluated systems, we selected GPT-4.1 as the LLM judge to avoid self-evaluation bias, a well-documented confound in LLM-as-judge frameworks whereby a model tends to favor outputs stylistically similar to its own generations. GPT-4.1 was chosen over alternative judge candidates for two reasons. First, its instruction-following reliability and structured output compliance made it well-suited for the fixed rubric scoring task, with fewer than 2\% of judge responses requiring re-querying due to malformed outputs. Second, its strong multilingual comprehension, particularly for morphologically rich languages, made it capable of detecting the honorific inconsistencies and structural violations central to our evaluation rubric qualities that a purely English-centric judge would likely miss.

\clearpage
\newpage
\vspace{1em}
\noindent

\vspace{2em} 

\vspace{2em} 

\noindent
\begin{minipage}{\textwidth}
  \centering

  \begin{tabular}{p{0.48\textwidth} p{0.48\textwidth}}
  \toprule
  \textbf{Before Fine-tuning(Generation and Register Failures)} & \textbf{After Fine-tuning} \\
  \midrule
  
  
  \textbf{Prompt: }{\fontspec{kalpurush.ttf}অগ্রিম ছুটির জন্য আবেদন}\\
  {\fontspec{kalpurush.ttf}...প্রাইভা, সাধারণত ১৫-১৭.০২.২০২৩ | 
  
শুখ: উচ্চকেন্দ্র বিবর্তন: ঢঙের ঘনমঞ্জর... 
  
ভারসাম্য: -২০°সে/-৪°ফা...
চীন চীন ভোচার: -৬০°সে/-৭৬°ফা... 
সালাম,

আপনাকে নম্রভাবে শুধুমাত্র 'অ' এর মাঝে বলা হচ্ছে।

আমি আপনার উপর আশা করি, তুমি প্রথমে খুব গৌণ ব্যক্তি কিনা? 

এই লিখিত অবেদনের সাথে আমি অগ্রিম ছুটি আবেদন করছি, যেমন:

1.   ফর্মের শীর্ষক - "অগ্রিম ছুটির জন্য আবেদন"
2.   ফর্মের ডেট - তারিখ (আপনি কোন তারিখ নির্ধারণ করেছেন?)
3.   ছুটির কারণ - আপনি কোন কারণে ছুটি আবেদন করছেন?
4.   স্থানীয় সংস্থা/প্রতিষ্ঠানের নাম - আপনি কোন স্থানীয় সংস্থার সাথে কাজ করছেন?
5.   ছুটির দৈর্ঘ্য - কত দিন পরিশোধ করবেন?

যদি তোমার এই অবেদন পূরণ করতে তৈরি করতে পারেন, তাহলে আমি আপনার উপর আশা করি.

(১) (২) (১)(০) চীন চীন ভারসাম্য:
-৮০°সে/-১১৩°ফা (-১১২°ফা) -১৫০°সে/-২৩৮°ফা...}

  & 

  
  \textbf{Prompt: }{\fontspec{kalpurush.ttf}অগ্রিম ছুটির জন্য আবেদন
  \vspace{0.1em}
  তারিখঃ ১৮/১০/২০২৪ খ্রিঃ \par
  বরাবর \par
  প্রধান শিক্ষক \par
  ছোলমাইদ হাই স্কুল এন্ড কলেজ \par
  ভাটারা, ঢাকা-১২১২ \par
  \textbf{বিষয়ঃ} অগ্রিম ছুটির জন্য প্রধান শিক্ষকের নিকট আবেদন। \par
  \vspace{0.1em}
  মহোদয়, \par
  সম্মানপূর্বক বিনীত নিবেদন এই যে, আমি আপনার বিদ্যালয়ের নবম শ্রেণীর একজন নিয়মিত ছাত্র। আগামী ২১/১০/২০২৪ থেকে ২৩/১০/২০২৪ তারিখ পর্যন্ত আমার বড় বোনের বিবাহ উপলক্ষে অগ্রিম ছুটির প্রয়োজন। উক্ত অনুষ্ঠানে আমাকে বড়দের কাজে সাহায্য করতে হবে তাই আমি বিদ্যালয়ে উপস্থিত হতে পারব না। \par
  \vspace{0.1em}
  অতএব মহোদয়ের নিকট আকুল আবেদন আপনাকে উল্লেখিত তিন দিনের ছুটি মঞ্জুর করে বাধিত করবেন। \par
  \vspace{0.1em}
  বিনীত, \par
  আপনার একান্ত অনুগত ছাত্র \par
  (আপনার নাম) \par
  শ্রেণী-নবম, রোল-১২ \par
  বিভাগ-বিজ্ঞান \par
  ছোলমাইদ হাই স্কুল এন্ড কলেজ \par
  ভাটারা, ঢাকা-১২১২ \par
  \vspace{0.1em}
  ব্যাপারের জন্য মহোদয়ের নিকট বিনীত আবেদন আপনার সার্বিক সুস্থতার জন্য অনুরোধ করছি যেন আপনি আমার অগ্রিম ছুটির আবেদনটি মানবিকভাবে বিবেচনা করেন।}
  \\
  \bottomrule
  \end{tabular}
  \captionof{table}{Model Output Before and After BLADE Fine-Tuning with LLaMA 3.2}
   \label{tab:spanning_comparison}
\end{minipage}

\begin{table*}[htbp]
\centering
\begin{tabular}{p{0.48\textwidth} p{0.48\textwidth}}
\toprule
\textbf{Before Fine-tuning (English)} & \textbf{After Fine-tuning (English)} \\
\midrule

\vspace{0pt} 
\textbf{Prompt:} Application for advance leave \par
\vspace{0.5em}
...Priva, usually 15-17.02.2023 |

Happiness: High center Evolution: Style's dense blossom...

Balance: -20°C/-4°F...
China China Voucher: -60°C/-76°F...
Greetings,

You [formal] are being humbly told only within the letter 'A'.

I hope upon you [formal], whether you [informal] are a very minor person first?

With this written mis-application, I am applying for advance leave, such as:

Form title - "Application for advance leave"

Form date - Date (you [formal] which date have determined?)

Reason for leave - you [formal] for what reason are applying for leave?

Name of local organization/institution - you [formal] with which local organization are working?

Length of leave - how many days will you [formal] disburse/repay?

If your [informal] this mis-application can prepare to fill up, then I hope upon you [formal].

(1) (2) (1)(0) China China Balance:
-80°C/-113°F (-112°F) -150°C/-238°F...

& 

\vspace{0pt} 
\textbf{Prompt:} Application for advance leave \par
\vspace{0.5em}
Date: 18/10/2024 AD \par
To \par
Headmaster \par
Solmaid High School and College \par
Vatara, Dhaka-1212 \par
\textbf{Subject:} Application to the Headmaster for advance leave. \par
\vspace{0.5em}
Sir, \par
With due respect, I state that I am a regular student of class nine in your school. I need advance leave from 21/10/2024 to 23/10/2024 on the occasion of my elder sister's marriage. I have to help the elders in the said ceremony, so I will not be able to attend school. \par
\vspace{0.5em}
Therefore, a fervent appeal to the Sir, that you would be kind enough to grant me the mentioned three days of leave. \par
\vspace{0.5em}
Sincerely, \par
Your most obedient student \par
(Your Name) \par
Class-Nine, Roll-12 \par
Department-Science \par
Solmaid High School and College \par
Vatara, Dhaka-1212 \par
\vspace{0.5em}
Humble appeal to Sir for the matter, requesting your overall well-being that you consider my advance leave application humanely. \\
\bottomrule
\end{tabular}
\caption{English-translation of Model Output Before and After BLADE Fine-Tuning with LLaMA 3.2.}
\label{tab:comparison_english}
\end{table*}




\begin{table*}[htbp]
\centering
\small
\begin{tabular}{lp{10cm}}
\toprule
\textbf{Source Website URL} & \\
\midrule
\url{https://allapplication.com} & \\
\url{https://probangla.com} & \\
\url{https://devlibrary.in} & \\
\url{https://bdsuggestion.com} & \\
\url{https://educarnation.com} & \\
\url{https://forms.portal.gov.bd} & \\
\url{https://www.somewhereinblog.net} & \\
\url{https://www.bissoy.com} & \\
\url{https://bengaliforum.com} & \\
\url{https://amaderforum.com} & \\
\url{https://www.scribd.com} & \\
\url{https://bn.wikisource.org} & \\
\url{https://archive.org} & \\
\url{https://bn.banglapedia.org} & \\
\url{https://www.jagonews24.com/education} & \\
\midrule
\multicolumn{2}{l}{\textbf{Books}}\\
\multicolumn{2}{p{14cm}}{\fontspec{kalpurush.ttf}{বাংলা ভাষার ব্যাকরণ (Bangla Bhashar Byakoron), বাংলা ব্যাকরণ ও নির্মিতি (Bangla Byakoron O Nirmiti)}}\\
\multicolumn{2}{p{14cm}}{\fontspec{kalpurush.ttf}{প্রমিত বাংলা ব্যাকরণ ও নির্মিতি (Promito Bangla Byakoron O Nirmiti), অক্ষরপত্র বাংলা ব্যাকরণ ও নির্মিতি}}\\
\multicolumn{2}{p{14cm}}{\fontspec{kalpurush.ttf}{উচ্চতর বাংলা ব্যাকরণ (Ucchotor Bangla Byakoron), বাঙ্গালা ব্যাকরণ (Bangala Byakoron)}}\\
\multicolumn{2}{p{14cm}}{\fontspec{kalpurush.ttf}{ভাষারীতি বাংলা ব্যাকরণ ও নির্মিতি (Bhashariti Bangla Byakoron O Nirmiti)}}\\
\multicolumn{2}{p{14cm}}{\fontspec{kalpurush.ttf}{শৈল্পিক বাংলা ব্যাকরণ ও নির্মিতি (Shoilpik Bangla Byakoron O Nirmiti)}}\\
\multicolumn{2}{p{14cm}}{\fontspec{kalpurush.ttf}{পাঞ্জেরী বাংলা ব্যাকরণ ও নির্মিতি (Panjeree Bangla Byakoron O Nirmiti)}}\\
\multicolumn{2}{p{14cm}}{\fontspec{kalpurush.ttf}{ভাষাজ্ঞান বাংলা ব্যাকরণ ও নির্মিতি (Bhashagyan Bangla Byakoron O Nirmiti)}}\\
\multicolumn{2}{p{14cm}}{\fontspec{kalpurush.ttf}{সহজ বাংলা ব্যাকরণ ও রচনা (Sahoj Bangla Byakoron O Rochona)}}\\
\multicolumn{2}{p{14cm}}{\fontspec{kalpurush.ttf}{বাংলা ২য় পত্র (ব্যাকরণ ও নির্মিতি) (Bangla 2y Potro - Byakoron O Nirmiti)}}\\
\multicolumn{2}{p{14cm}}{\fontspec{kalpurush.ttf}{এইচএসসি বাংলা দ্বিতীয় পত্র, মাস্টারবুক (HSC Bangla Ditiyo Potro Masterbook)}}\\
\multicolumn{2}{p{14cm}}{\fontspec{kalpurush.ttf}{অভিনব বাংলা ব্যাকরণ (Abhinob Bangla Byakoron), বাংলা ব্যাকরণ প্রসঙ্গ (Bangla Byakoron Prosongo)}}\\
\bottomrule
\end{tabular}
\caption{List of Source Websites \& Books for Bangla Language Content}
\label{tab:website_sources_url_only}
\end{table*}
\begin{table*}[htbp]
\centering

\begin{tabular}{|p{3cm}|p{6cm}|p{6cm}|}
\hline
\textbf{Category} & \textbf{Before Fine-tuning} & \textbf{After Fine-tuning} \\
\hline
\textbf{Input Prompt} & {\fontspec{kalpurush.ttf}অগ্রিম ছুটির জন্য আবেদন} & {\fontspec{kalpurush.ttf}অগ্রিম ছুটির জন্য আবেদন} \\
\hline
\textbf{Language Quality} & Mostly gibberish and inconsistent text, mixed English–Bengali–random words. & Fluent and grammatically correct Bengali throughout. \\
\hline
\textbf{Structure} & No logical structure, random temperature-like values and disjointed phrases. & Proper formal letter structure with date, recipient, subject, body, and closing. \\
\hline
\textbf{Context Understanding} & Misinterprets the context — output unrelated to “leave application”. & Correctly identifies the task and generates a leave application letter. \\
\hline
\textbf{Relevance} & Irrelevant text (weather data, nonsense symbols). & Highly relevant, specific to “advance leave request”. \\
\hline
\textbf{Tone and Formality} & Nonsensical tone, lacks coherence or formality. & Formal, polite tone appropriate for a student–teacher letter. \\
\hline
\textbf{Output Length} & Short, disorganized, meaningless lines. & Well-developed paragraph-level letter, complete and coherent. \\
\hline
\textbf{Overall Accuracy} & 1/10 – completely wrong. & 9.5/10 – semantically, structurally, and linguistically accurate. \\
\hline
\end{tabular}
\caption{Comparison of LLaMA 3.2 Model Performance Before and After Fine-tuning}
\label{tab:llama32_comparison}
\end{table*}

\end{document}